\documentclass[letterpaper, 10pt, conference]{IEEEtran}

\IEEEoverridecommandlockouts                        

\usepackage{graphicx} 
\usepackage{amsmath} 
\usepackage{amsfonts}       
\usepackage{nicefrac}       
\usepackage{microtype}      
\usepackage{subfig}			
\usepackage{tikz}
\usepackage{adjustbox}
\usepackage{booktabs}	
\usepackage{multirow}
\usepackage{url}

\newcommand{\fillbox}[3]
{\bgroup
  \dimen1=#1\relax
  \dimen2=#2\relax
  \sbox0{\includegraphics[width=#1]{#3}}%
  \ifdim\ht0>\dimen2
    \dimen0=\dimexpr \ht0-\dimen2\relax
    \adjustbox{clip=true, trim=0pt 0.5\dimen0 0pt 0.5\dimen0}{\usebox0}%
  \else
    \sbox0{\includegraphics[height=#2]{#3}}%
    \ifdim\wd0>\dimen1
      \dimen0=\dimexpr \wd0-\dimen1\relax
      \adjustbox{clip=true, trim=0.5\dimen0 0pt 0.5\dimen0 0pt}{\usebox0}%
    \else
      \usebox0
    \fi
  \fi
\egroup}

\newcommand{\ie}{{\em i.e.}}
\newcommand{\eg}{{\em e.g.}}

\DeclareMathOperator*{\argmin}{argmin}

\title{\LARGE \bf
Learning Action Duration and Synergy in Task Planning for Human-Robot Collaboration
}

\author{Samuele Sandrini$^{1}$, Marco Faroni$^{1}$, Nicola Pedrocchi$^{1}$
\thanks{This work was partially supported by ShareWork project (H2020, European Commission -- G.A. 820807).}
\thanks{$^{1}$ STIIMA-CNR - Institute of Intelligent Industrial Technologies and Systems, National Research Council of Italy 
{\tt\small \{name.surname\}@stiima.cnr.it}}%
}

\begin{document}

\maketitle
\thispagestyle{empty}
\pagestyle{empty}

\begin{abstract}
A good estimation of the actions' cost is key in task planning for human-robot collaboration.
The duration of an action depends on agents' capabilities and the correlation between actions performed simultaneously by the human and the robot.
This paper proposes an approach to learning actions' costs and coupling between actions executed concurrently by humans and robots.
We leverage the information from past executions to learn the average duration of each action and a synergy coefficient representing the effect of an action performed by the human on the duration of the action performed by the robot (and vice versa).
We implement the proposed method in a simulated scenario where both agents can access the same area simultaneously. 
Safety measures require the robot to slow down when the human is close, denoting a bad synergy of tasks operating in the same area.
We show that our approach can learn such bad couplings so that a task planner can leverage this information to find better plans.
\end{abstract}


\section{Introduction}\label{sec: intro}

Human-Robot Collaboration (HRC) often requires the system to make decisions based on human users' observed or predicted behavior.
In such a context, planning and allocating tasks to the human and robot agents are two complex problems, even for tasks composed of a few activities. Two modelling issues are essential. First, human behavior is intrinsically unpredictable and partially uncontrollable  \cite{PELLEGRINELLI20171}. Second, the coupling between the human and the robotic agents is characterized by a large variability. 
For example, the feasibility and the duration of an action may vary because of the interference between the human and the robot (\emph{e.g.}, safety stops of the robot or even due to path-switch in motion re-planning \cite{Tonola_ROMAN2021}).

%

In recent years, task planning and allocation problems for HRC have been investigated.
Existing works tried to model human preferences explicitly and ergonomics into planning \cite{T-CYB-human-agent-collaboration}.
For example, \cite{Alami-HATP, Alami-Lallement-HATP} proposed a hierarchical agent-based task planner, where complex tasks can be decomposed into simpler actions. This approach can improve the collaborative experience by considering human preferences as social costs \cite{Alami-using-human-knowledge, Alami-dealing-with-online-human}. 
Manufacturing-oriented works focus on process throughput by minimizing the expected duration \cite{Lippi2021} or planning contingent plans to reduce process errors \cite{Rosell-knowledge-oriented-tamp, Rosell-Applied-Sciences}.
A promising approach to deal with uncertainty on the task duration is timeline-based planning \cite{cialdeaorlandini2015}.
Timeline-based planning explicitly models the duration variability of tasks and finds plans that are robust with respect to it.
Timeline-based planning was used in HRC with pre-computed robot motions \cite{Pellegrinelli2017}, and online motion planning \cite{Faroni_ROMAN2020}, demonstrating robust plans allow for less frequent re-planning and shorter average task duration.

Most planning-based methods assume a guess of the action cost (e.g., the duration) is available from a domain expert.
If this information is unreliable, the task planner reasons on wrong assumptions so that the resulting plans become sub-optimal or even infeasible (leading to frequent re-planning when using a motion re-planner such as \cite{Tonola_ROMAN2021}).
Moreover, the duration guess does not consider possible couplings between the tasks executed concurrently by the human and the robot.
Therefore, the task planner neglects each agent's positive or negative effects on the others.
For example, if concurrent tasks require the human and the robot to work in the same area, the robot would probably slow down because of safety.
If the task planner knew this effect, it could favour pairs of tasks that avoid robot safety slowdowns.

In this work, we propose a method to learn the expected duration of a joint plan.
We leverage a minimum-time formulation of the task planning/allocation problem for HRC.
Then, we estimate the expected duration of each task from previous executions and learn a synergy coefficient that represents the effect of one task over the other. 
A synergy coefficient greater than one means that the human task causes a slow down of the robot task.
We show that it is possible to cast the estimation of the synergy coefficients into a set of linear regression problems (one for each task).
The resulting synergy coefficients can be exploited in task planning and allocation method to select advantageous task couplings.
The proposed approach can be used offline -- to obtain a guess of the task duration and synergies -- or iteratively as the number of executions grows to refine the initial guess over time.
We demonstrate the proposed approach in an HRC scenario where a human and a collaborative robot shall perform a sequence of pick-and-place operations.
Experiments show that pairs of tasks that drive the nearby agents lead to high synergy coefficients, recognizing favorable and unfavourable pairs.

The paper is organized as follows.
Section \ref{sec: preliminaries} defines the task planning problem for HRC systems.
Section \ref{sec: methodology} builds on that model to formulate the task expected duration in terms of average duration and synergy coefficients. Then, it casts the estimation of such coefficients into a set of linear regression problems.
Section \ref{sec: framework-architecture} describes the software architecture to collect and process the data from task executions.
Section \ref{sec: experiments} applies the proposed methodology to the manufacturing example and shows that the resulting coefficients reflect the good and bad coupling effects of robot safety stops.
Finally, conclusions and future works are discussed in Section \ref{sec: conclusions}.

\section{Preliminaries}\label{sec: preliminaries}

\subsection{Task Planning Problem}
\label{subsec: task-planning-problem}
A task planning problem can be formalized as an optimization problem. Given a set $\{H, R\}$ of agents, \ie, human and robot, and a set of Tasks $\mathcal{T} =\{\tau_i\}$, the objective of the problem is to obtain a task plan and assignment $\pi$ that minimizes the duration of the process.

We denote a task by a tuple $\tau = (l, d, t)$, in which:
\begin{itemize}
    \item $l \in \{H,R\}$ is an assignment variable that specifies which agent can perform the specific task, \ie, human, or robot;
    \item $d \in \mathbb{R}^{+}$ is a guess of the task duration;
    \item $t = \bigl [t^{\text{start}} \; ; \; t^{\text{end}} \bigl]$ is an interval with endpoints corresponding to the start and end time.
\end{itemize}

We refer to $\mathcal{T}^H$ and $\mathcal{T}^R$ as the subsets of $\mathcal{T}$ such that $l = H$ and $l = R$, respectively. 

In this context, it is possible to introduce a binary assignment variable that defines the allocation of $\tau_i \in \mathcal{T}$ to the robot ($a^R_{i}$) or to the human ($a^H_{i}$):
\begin{equation}
\begin{aligned}
 a^R_{i} &= \left \{
 \begin{aligned}
  &1, && \text{if task $\tau_i$ is assigned to the robot} \\
  &0, && \text{otherwise}
 \end{aligned} \right.\\
 a^H_{i} &= \left \{
 \begin{aligned}
  &1, && \text{if task $\tau_i$ is assigned to the human }\\
  &0, && \text{otherwise.}
 \end{aligned}\right.
\end{aligned}
\end{equation} 
The duration of a plan $\pi$ is the maximum between the duration of the robot's and the human's plan, denoted by $ d^H_\pi$ and $ d^R_\pi$ respectively.
The duration of each agent's plan can be calculated as:
\begin{equation}
\label{eq-agent-plan-duration}
    \begin{aligned}
    d^H_\pi &= \sum_{i} d^H_i a^H_i \\
    d^R_\pi &= \sum_{i} d^R_i a^R_i 
    \end{aligned}
\end{equation}
By defining a cost function $J$ that represents the duration of a plan $\pi$ as:
\begin{equation}
\label{eq-duration-function}
J = \max \{ d_\pi^H, d_\pi^R \}
\end{equation}
the optimal plan $\pi^*$ is:
\begin{equation}
\label{optimal-problem-solution}
	\pi^* = \argmin_\pi {J}
\end{equation}
subject to the constraint that $\tau_i \in \mathcal{T}$ can only be assigned during $\pi$:
\begin{equation}
 a^H_i+a^R_i=1 \quad\forall\ i \;\; \mbox{s.t.}\;\;\tau_i \in \mathcal{T}
\end{equation}
and assignment, temporal, and causal constraints owed to the process requirements\footnote{If \eqref{optimal-problem-solution} only involves assignment constraints, the problem is a task allocation; if causal and/or temporal constraints are considered, the problem becomes a task planning\&scheduling problem.}.

\section{Methodology}\label{sec: methodology}

The task planning problem assumes that a guess of the task duration exists. A domain expert usually provides this guess. 
However, this information is often unreliable and comes with a sizeable unmodeled uncertainty.
Moreover, it does not account for the effect of the task performed by other agents.
Indeed, the task duration is a function of the tasks executed simultaneously by the other agent, i.e.:
    \[
    \begin{aligned}
    d^R_i &= d^R_i(\tau^H_j)  \;\; \forall\ j \;\; \mbox{s.t.}\;\; \tau^H_j \;\; \mbox{:} \;\; t^H_j \cap  t^R_i \neq \emptyset\\
    d^H_i &= d^H_i(\tau^R_j)  \;\; \forall\ j \;\; \mbox{s.t.}\;\; \tau^R_j \;\; \mbox{:} \;\; t^R_j \cap t^H_i \neq \emptyset
    \end{aligned}
    \]
where $\tau^H_j$ and $\tau^R_j$ are tasks assigned to humans and robots.
If the coupling between tasks is not modelled, the optimal task  \eqref{optimal-problem-solution} may be unreliable when executed on the real-world system.
To overcome this problem, we extend the formulation given in \ref{subsec: task-planning-problem} to include a synergy coefficient in the task duration.

\subsection{Task Planning Problem with Explicit Task Coupling}

For each couple of task indices $(i, j)$, a synergy term is introduced for each agent and denoted with $s^R_{i, j}$ for the robot agent and $s^H_{i, j}$ for the human agent. 
The synergy term denotes the increment of the duration of task $\tau_i$ when executed by the robot while the human is executing task $\tau_j$.
We define this coefficient as:
\label{subsec: synergy-estimation}
\begin{equation}
    \label{synergy-term}
    s^R_{i, j}=\frac{d^R_{i, j}}{\hat{d}^R_i}
\end{equation}
where $d^R_{i, j}$ is the expected duration of task $\tau_i$ when the human executes $\tau_j$ and $\hat{d}^R_i$ is the expected value of the duration of $\tau_i^R$ for all concurrent tasks $\tau_j^H$. 
Thus, the duration of a plan $\pi$ becomes:
\begin{equation}
    \label{plan-duration-synergy}
    \begin{aligned}
    d^R_\pi &= \sum_{i} \hat{d}^R_i a^R_i \biggl( \sum_{j} s^R_{i, j} \delta^R_{i, j} a^H_j \biggl)
    \end{aligned}
\end{equation}
where $\delta^R_{i, j}$ represents the ratio of the overlapping time between  $\tau^R_i$ and $\tau^H_j$ with respect to the duration of the task $\tau^R_i$, thus defined as:
\begin{equation}
    \label{overlapping-ratio}
    \delta^R_{i,j} = \frac{D( t^H_j \cap t^R_i )}{D(t^R_i)}
\end{equation}
and $D$ is a function that calculates the duration of a temporal interval $t=[t^{start}, t^{end}]$: 
\begin{equation}
 D(t) = \left \{
 \begin{aligned}
  &t^{end} - t^{start}, && \text{if $t \neq \emptyset$} \\
  &0, && \text{if $t = \emptyset$ }
 \end{aligned} \right .
\end{equation}
Equations \eqref{synergy-term}, \eqref{plan-duration-synergy}, \eqref{overlapping-ratio} can be rewritten for the human agent as:

\begin{align}
        s^H_{i, j} &= \frac{d^H_{i, j}}{\hat{d}^H_i}                        \\
        \delta^H_{i,j} &= \frac{D( t^R_j \cap t^H_i )}{D(t^H_i)}        \\
        \label{plan-duration-synergy-human}
        d^H_\pi &= \sum_{i} \hat{d}^H_i a^H_i \biggl( \sum_{j} s^H_{i, j}\delta^H_{i, j} a^R_j \biggl) 
\end{align}

Using \eqref{plan-duration-synergy} and \eqref{plan-duration-synergy-human} in \eqref{eq-duration-function}, a task planning can minimize the process duration taking into account the coupling effect between concurrent tasks.

\subsection{Synergy and duration estimation}
\label{subsec: synergy-estimation}
We estimate both task duration and synergy coefficients from experience.
Given $n$ executions of a task $\tau_i$, we approximate its duration with its expected value:

\begin{equation}
    \begin{aligned}
        \hat{d}^R_i &\approx \mathbb{E}\biggl[d^R_i\rvert_{k}\biggl]   \quad\forall\ k=\{1, \dots, n\}     \\
        \hat{d}^H_i &\approx \mathbb{E}\biggl[d^H_i\rvert_{k}\biggl]   \quad\forall\ k=\{1, \dots, n\}     \\
    \end{aligned}
\end{equation}
where $d^R_i\rvert_{k}$ and $d^H_i\rvert_{k}$ are the task duration measured in execution $k$.

Furthermore, it is possible to formalize the problem of estimating task synergy coefficients as a least-square regression problem. For each sample $k$: 
\begin{equation}
    \label{regression-row}
	D(t^H_i) \Big \rvert_{k} = \sum_{j=1}^{m} \delta^H_{i, j} \Big \rvert_{k} s^R_{i, j} d^R_i + T_{\text{idle}}\Big \rvert_{k}
\end{equation}
where $T_{\text{idle}}$ is the time when the robot is not assigned to any task during $\tau^H_i$ and $m=|\mathcal{T}^R|$ .
Equation \eqref{regression-row} can be written in matrix form as:
\begin{equation}
\label{regression-problem}
    \mathbf{D_{i}^H} = \mathbf{R^R} \mathbf{S^R_i} + \mathbf{T_{\text{idle}}} 
\end{equation}
where:
\begin{itemize}
    \item $\mathbf{D_{i}^H} \in \mathbb{R}^{n \times 1}$ is a column vector containing the human execution task duration $D(t^H_i)\Big \rvert_{k}$;
    \item $\mathbf{R^R} \in \mathbb{R}^{n \times m}$ is the regression matrix, which takes the following form:
    \begin{equation}
    \label{regression-matrix}
        \mathbf{R^R} =
                \begin{bmatrix}
                 \delta^H_{i, 1} \Big \rvert_{1} & \cdots &  \delta^H_{i, m} \Big \rvert_{1} \\
                \vdots     & \ddots & \vdots                 \\
                 \delta^H_{i, 1} \Big \rvert_{k} & \cdots &  \delta^H_{i, m} \Big \rvert_{k} \\
                \vdots     & \ddots & \vdots                 \\
                          
                \delta^H_{i, 1} \Big \rvert_{n} & \cdots &  \delta^H_{i, m} \Big \rvert_{n} \\
            \end{bmatrix}d^R_i
        \end{equation}
    \item $\mathbf{S^R_i} \in \mathbb{R}^{m \times 1}$ is a column vector containing the synergy coefficients to be estimated:
    \begin{equation}
    \label{synergy-vector}
        \mathbf{S^R_i}=
    	\begin{bmatrix}
    	    s^R_{i, 1} &
    	    s^R_{i, 2} &
    	    \cdots &
    	    s^R_{i, m} 
    	\end{bmatrix}^T
    \end{equation}
    \item $\mathbf{T_{\text{idle}}} \in \mathbb{R}^{n \times 1}$ is a column vector containing the robot idle times $T_{\text{idle}}\Big \rvert_{k}$.
\end{itemize}

In conclusion, the solution of the regression problem in \eqref{regression-problem} is obtained from the following:
\begin{equation}
\label{regression-solution}
	\mathbf{S^R_i} = \Bigl(\mathbf{R^R} ^T \mathbf{R^R} \Bigl)^{-1} \mathbf{R^R} ^T \Bigl[\mathbf{D_{i}^H} - \mathbf{T_{\text{idle}}}\Bigl].
\end{equation}
and using the same nomenclature for the human agent:
\begin{equation}
\label{regression-solution-human}
	\mathbf{S^H_i} = \Bigl(\mathbf{R^H} ^T \mathbf{R^H} \Bigl)^{-1} \mathbf{R^H} ^T \Bigl[\mathbf{D_{i}^R} - \mathbf{T_{\text{idle}}}\Bigl].
\end{equation}

\section{Proposed framework architecture}
\label{sec: framework-architecture}

\begin{figure}[t]
  \centering
  \includegraphics[width=\columnwidth]{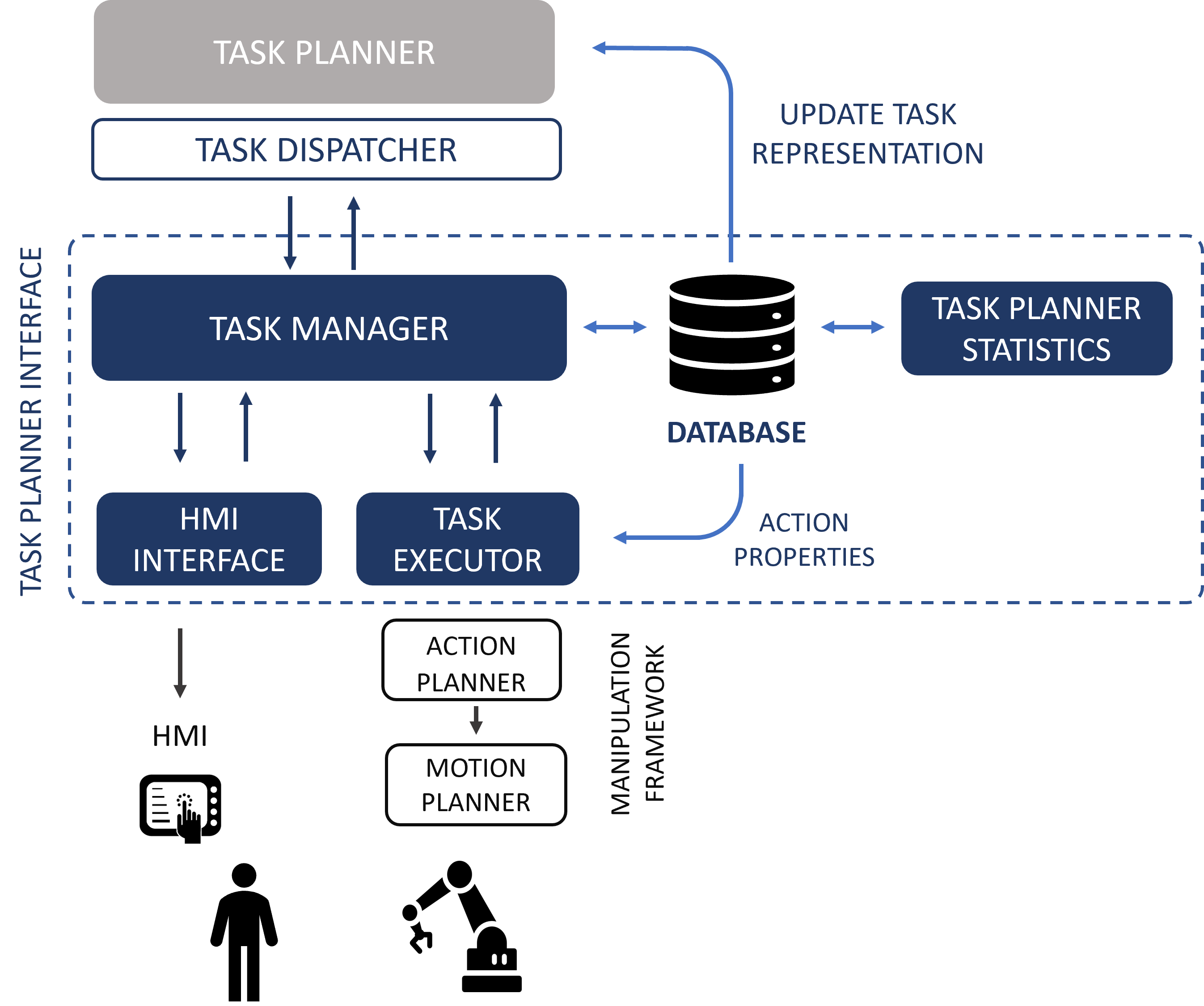}
  \caption{Proposed framework architecture.}
  \label{fig: framework}
\end{figure}

Figure \ref{fig: framework} shows the software architecture developed to measure and estimates action duration and synergies for real-world deployment of our approach.

\begin{figure*}[t]
  \centering
  \includegraphics[width=0.9\textwidth]{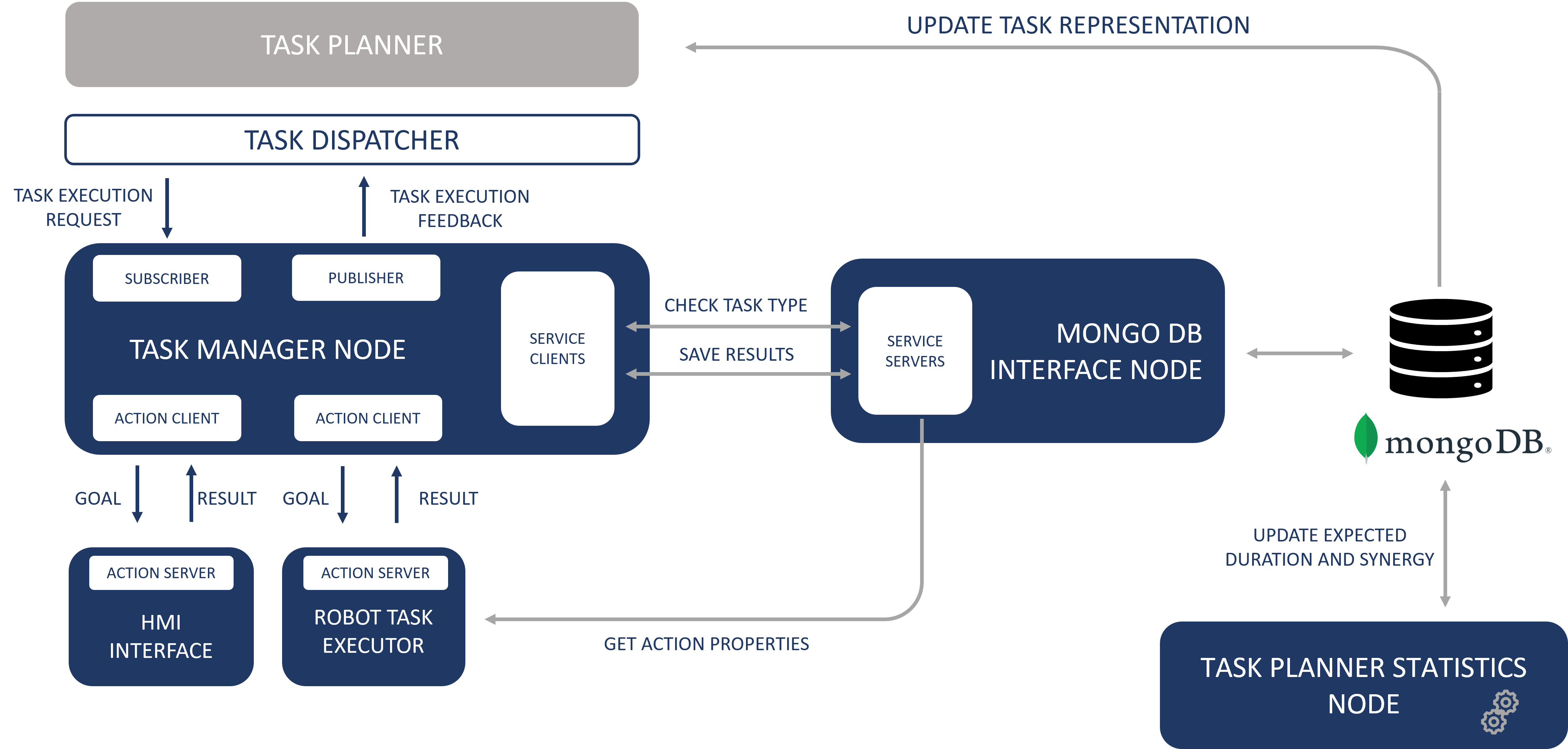}
  \caption{Software Implementation.}
  \label{fig: sw_implementation}
\end{figure*}

The \emph{Task planner} is at the highest level and has a symbolic knowledge of the tasks to execute. Based on the temporal constraints between tasks (\ie, precedence) and a-priori durations information, it computes the optimized plan, the scheduling, and the assignment of tasks to agents. This information is used by the \emph{dispatcher}, which receives the plan and sends the request at a lower level; finally, it waits for the result to proceed with the remaining part of the plan. 

The \emph{task planner interface} module is at the middle layer. The first module component is the \emph{task manager}, which receives the task request from the dispatcher and queries the database to check for properties associated with the requested symbolic task. In a positive case, the \emph{task manager} sends the task execution request to the task executor of the agent specified by the task planner. 
When feedback comes from the underlying layer, the \emph{task manager} module interacts with the \emph{database} to update the task results.
The core of the \emph{task planner interface} is a database representing the high-level updated Knowledge-Base. 
The database stores information used at different levels: the definition of high-level task properties, run-time information of duration execution, statistical information of expected duration, and synergy of concurrent tasks between agents.

At the lowest level, there is the single-agent \emph{task executor}, which converts symbolic information associated with tasks into geometric targets. First, it interacts with the database to retrieve the action type (\eg, pick, place, go to) and the symbolic goal associated with the task. 
This layer has a geometric knowledge of the system and translates the symbolic goal into a sequence of robot movements. 
To do so, it knows the locations of symbolic goals, integrates a motion planner, and may include an action planner such as in \cite{Faroni_ROMAN2020}. 
Once the task execution is finished, the feedback is sent to the layer above.

The \emph{task manager} sends the task request to an \emph{HMI interface}, which communicates to the HMI the information of the task to be executed and waits for the confirmation of execution. 
Then, it sends the feedback to the \emph{task manager}.

The advantage of this architecture is that, through the presence of the database, it is possible to keep the tasks' representation up-to-date, on which the high-level planning is based. 
So, it is possible to update actions' costs based on the experience of executions.
For this purpose, the \emph{task planner statistics} module calculates the estimated duration and task synergy as described in Section \ref{subsec: synergy-estimation}.

\subsection{Software implementation}
\label{subsec: software-implementation}

The functional architecture presented in Section \ref{sec: framework-architecture} was developed as a hybrid Python/C++ library based on ROS \cite{ros-paper}.

Software development focused on the task planner interface level. 
The crossroads component is the \emph{task service manager}. As shown in Figure \ref{fig: sw_implementation}, this node exploits the Publisher-Subscriber communication to receive the task execution request from the dispatcher and then send back the feedback.

The communication with the database occurs through an interface node that acts as a service server for the task service manager: a service checks for the existence of task properties associated with the symbolic task, and another service saves the execution results from the layer below. 

Communication with the underlying layer, \ie, \emph{task execution} nodes, takes place according to the client-server action model. The task service manager sends a goal containing a symbolic task name of the task to be executed and waits for the results, \eg, duration, task type, and agent information.

The database is composed of different collections, each dedicated to storing different information:
\begin{itemize}
    \item \emph{task properties}: for each task, it stores a symbolic identifier, the type of action associated with it (\eg, pick, place, go to), a textual description, the agents that can execute it, and the symbolic goals associated with it.
    \item \emph{Task results}: it is updated in real-time with information from the lower layers about the execution time and the success or failure of the task.
    \item \emph{Task duration}: it contains the information of expected durations and standard deviation associated with each symbolic task.
    \item \emph{Task synergy}: for each pair of tasks, it contains the result of the estimation described in Section \ref{subsec: synergy-estimation}. 
\end{itemize}
Notice that both \emph{Task duration} and \emph{Task synergy} can be updated synchronously or asynchronously. 

In this framework, MongoDB was used as the database \cite{Mongo_VS_MySql}. In addition, PyMongo library was used for interaction with MongoDB database in the so-called \emph{Mongo-DB interface node} shown in Figure \ref{fig: sw_implementation}.

At the lowest level, the \emph{task-execution} node interacts with the database interface to retrieve the geometric goal associated with the task. 
It acts as a client of the manipulation framework \cite{Manipulation}, to which it sends requests, specifying the type of action to be performed. 
The manipulation framework defines the sequence of movements required to perform the requested action, solves the corresponding motion planning problem, and finally executes the action.

Finally, the \emph{task planner statistics node} is independent of the framework information flow and provides ROS-Services to update the \emph{task duration} and the \emph{task synergy} collections and make statistical charts. These services can be called synchronously or asynchronously concerning the information flow of the framework.

\section{Experiments}\label{sec: experiments}

\begin{figure}[t]
  \centering
  \includegraphics[width=\columnwidth]{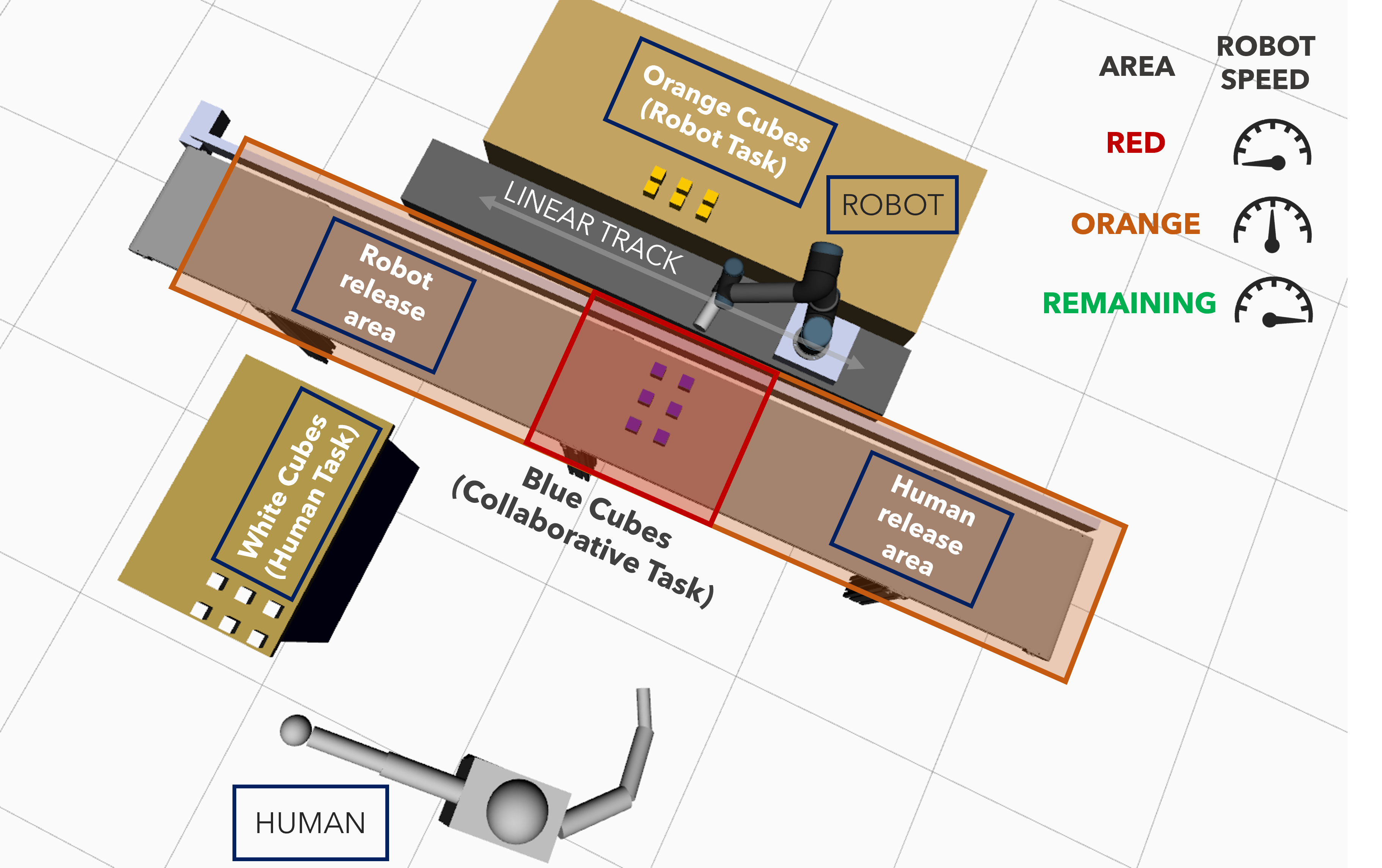}
  \caption{Simulated collaborative workcell}
  \label{fig: collaborative-workcell}
\end{figure}
We test our approach in a simulated case study of a collaborative workcell (Figure \ref{fig: collaborative-workcell}) composed of a collaborative robot, UR5, mounted on a linear axis and a human operator. The agents share a portion of the workspace, defining a collaborative workspace.
There are six white cubes on the work table accessible only by the human operator, six orange cubes on the work table accessible by the robot, and six blue cubes within the collaborative work area accessible by both agents. Thus, humans can handle only white and blue objects, while robots can handle only orange and blue objects.

Each agent has a dedicated object release area, as illustrated in Figure \ref{fig: collaborative-workcell}.  
The process goal is that: (i) the robot picks four orange and two blue boxes by placing them in its release area; (ii) the human picks four white and two blue boxes by arranging them in its release area. 
Therefore, the set of tasks that the human and the robot can perform is defined by:
\begin{multline}
\label{task-domain-human}
    \mathcal{T}^H=\{\text{Pick Orange Box}, \text{Place Orange Box},\\
    \text{Pick Blue Box (H)}, \text{Place Blue Box (H)}\}
\end{multline}
\begin{multline}
\label{task-domain-robot}
      \mathcal{T}^R=\{\text{Pick White Box},
      \text{Place White Box},\\
      \text{Pick Blue Box (R)}, \text{Place Blue Box (R)}\}
\end{multline}
where $\mathcal{T} = \mathcal{T}^H \cup \mathcal{T}^R$.

The process requires access to the same work area, especially during a simultaneous execution of the \emph{Pick Blue Box} and \emph{Place Blue Box} tasks.
The symbolic goal associated with each task does not refer to a specific pick or place slot. Thus, the choice of the best slot associated with a symbolic goal is performed by the action planner of the manipulation framework \cite{Manipulation} at runtime by solving a multi-goal motion planning problem.

The ISO/TS 15066 \cite{ISOTS15066} is applied to ensure safety. Three collaborative areas are defined. If any human body part enters inside the red area, the robot is stopped. If the human enters the orange area, the robot moves at 50 \% of its nominal speed. Finally, if the human operator moves into the remaining area, the robot is free to move at its nominal speed.

\subsection{Results}
In order to estimate the synergy coefficients between tasks, 50 plans are randomly generated respecting the assignment constraints imposed by the agent's domains \eqref{task-domain-human} and \eqref{task-domain-robot} and precedence constraints between a pick and place tasks.
A simulation run is performed for each recipe, using the framework proposed in Section \ref{sec: framework-architecture}: tasks are executed, and the results are saved to the database.

\begin{figure}[t]
  \centering
  \includegraphics[trim={0 0 0 1cm},clip,width=\columnwidth]{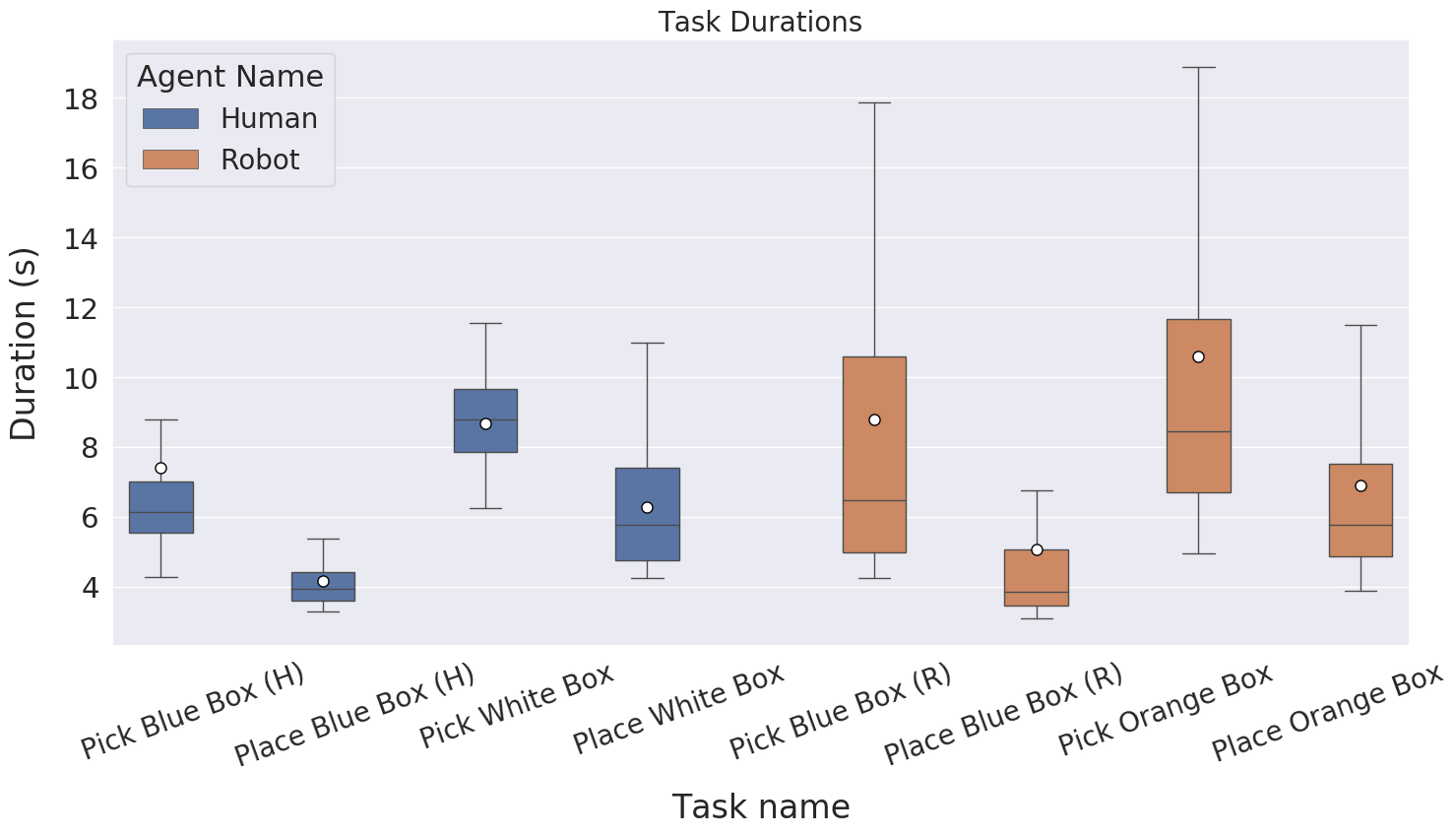}
  \caption{Task durations grouped by agent, white circles denote the  mean.}
  \label{fig: task-duration}
\end{figure}

We exploit the \emph{task planner statistics} node to calculate the expected duration and standard deviation for each task (grouped by type and agent). Results are shown in Figure \ref{fig: task-duration}.

For each type of task, outliers are removed using an isolation forest algorithm to reduce noise from the task results before synergy coefficients estimation \cite{scikit-learn}. 

\begin{figure}[t]
  \centering
  \includegraphics[trim={0 0 0 1cm},clip,width=\columnwidth]{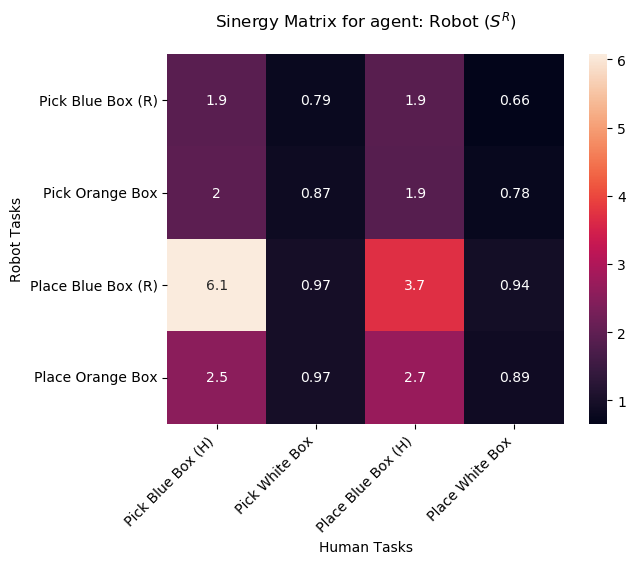}
  \caption{Robot task Synergy Matrix (each row corresponds to the vector $S^R_i$ obtained from \eqref{synergy-vector})}
  \label{fig: synergy-matrix}
\end{figure}

We use the estimated duration of each task and saved task results to apply the regression described in \eqref{regression-problem} and obtain the synergy coefficients estimation between robot and human tasks. 
We obtain the column vector in \eqref{synergy-vector} by applying the regression for each task. Repeating this procedure for each task, we obtain the synergy matrix shown in Figure \ref{fig: synergy-matrix} (a heatmap graphically highlighting the positive or negative synergy between concurrent tasks). 
The heatmap shows the synergy coefficients of the robot agent task versus the human tasks. The tasks of the robot \eqref{task-domain-robot} are placed in the rows of the heatmap, and on the columns are the human tasks \eqref{task-domain-human}. 
An $s^R_{i, j}$ element of the heatmap reports the duration increment of the robot's task $\tau_i$ when simultaneous with the human task $\tau_j$.

If the value $s^R_{i, j}$ is greater than 1, it means that task $\tau_i$ performed by the robot is penalized when it is in parallel with task $\tau_j$ performed by the human. Vice versa, if $s^R_{i, j}$ is smaller than 1, it means that the coupling between tasks $\tau_i$ and $\tau_j$ is advantageous in terms of execution time.

The heatmap analysis shows that all the robot tasks are penalized when the human simultaneously performs the \emph{Pick Blue Box} and \emph{Place Blue Box} tasks. Indeed, when the human operator executes \emph{Pick Blue Box}, he accesses a \emph{red area} close to the robot, causing the robot to stop. Conversely, the robot tasks do not slow down or even improve performance when they are executed in parallel with the human tasks \emph{Pick White Box} (which does not cause an access to the collaborative area), and \emph{Place White Box}.

Figure 6 also shows the synergy coefficients and their standard deviation estimates by the regression to validate the statistical significance of the estimate.

\begin{figure}[t]
  \centering
  \includegraphics[trim={0 0 0 0.7cm},clip,width=\columnwidth]{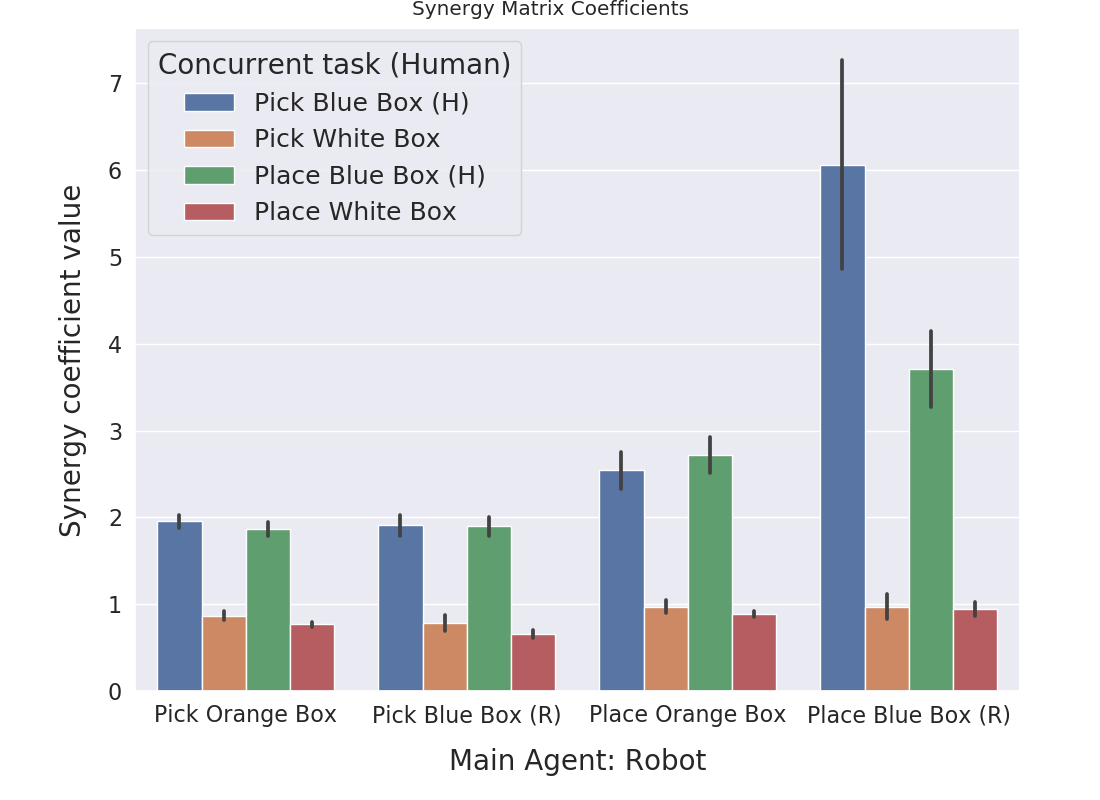}
  \caption{Synergy coefficients and their standard deviation}
  \label{fig: synergy-coefficients}
\end{figure}

\section{Conclusions and future works}\label{sec: conclusions}

In this paper, a synergy term that takes into account the coupling between agent tasks is defined. The synergy coefficient is contextualized in a generic task planning problem by including it in the duration cost function.

We validated the approach in a simple HRC scenario where the positive of negative couplings were intuitive in order to show that the proposed method can learn the expected good or bad synergy of pairs of tasks.

Future works will focus on integrating the estimated synergy term in a task planner to demonstrate the reduction of task plan duration in complex scenarios.

Alternative strategies will be compared to access the advantages of the proposed approach.

\bibliographystyle{IEEEtran}
\bibliography{bib, new_bib, reference_stiima, references}

\end{document}